\documentclass{article}
\usepackage{ijcai09}

\usepackage{times}

\usepackage{latexsym} 
\usepackage{amsthm} 

\usepackage{graphicx}
\usepackage{xspace}

\newcommand{\cb}[2][CB]
{$^{\fboxsep=1pt\fbox{\tiny #1}}$%
\marginpar{\fbox{\parbox[t]{\marginparwidth}{\tiny #2}}}}

\newtheorem{mytheorem}{Theorem}
\newtheorem{mylemma}{Lemma}
\newcommand{\myproof}{\noindent {\bf Proof:\ \ }}

\newcommand{\myqed}{\mbox{$\Box$}}
\newtheorem{myexample}{Example}{\bf}{\it}

\newcommand{\constraint}[1]{\mbox{\sc #1}}
\newcommand{\alldiff}{\constraint{All-Different}\xspace}
\newcommand{\regular}{\constraint{Regular}\xspace}
\newcommand{\grammar}{\constraint{Cfg}\xspace}

\newcommand{\gcc}{\constraint{GCC}\xspace}
\newcommand{\egcc}{\constraint{eGCC}\xspace}

\newcommand{\nvalue}{\constraint{NValue}\xspace}
\newcommand{\permutation}{\constraint{Permutation}\xspace}
\newcommand{\mypermutation}{\constraint{Same}\xspace}

\newcommand{\usedby}{\constraint{Used-By}\xspace}
\newcommand{\uses}{\constraint{Uses}\xspace}

\newcommand{\common}{\constraint{Common}\xspace}

\newcommand{\sequence}{\constraint{Sequence}\xspace}
\newcommand{\precedence}{\constraint{Precedence}\xspace}

\newcommand{\roots}{\constraint{Roots}\xspace}
\newcommand{\range}{\constraint{Range}\xspace}

\newcommand{\myOmit}[1]{}

\title{Decompositions of All Different, Global Cardinality 
and Related Constraints}
\author{Christian Bessiere\thanks{Supported by the project  ANR-06-BLAN-0383-02.}
\\
LIRMM, CNRS\\
Montpellier\\
bessiere@lirmm.fr
\And
George Katsirelos\thanks{NICTA is funded by the Australian Government 
through the Department of Broadband, Communications and the Digital Economy and the Australian Research Council.}\\ 
NICTA, Sydney\\
gkatsi@gmail.com
\And
Nina Narodytska$\mbox{}^{\dagger}$\\
NICTA and UNSW\\
Sydney\\
ninan@cse.unsw.edu.au
\And
Claude-Guy Quimper\\
{EPM}, Montr\'{e}al\\
cquimper@gmail.com
\And
Toby Walsh$\mbox{}^{\dagger}$\\
NICTA and UNSW\\
Sydney\\
toby.walsh@nicta.com.au
}

\begin{document}

\maketitle

\begin{abstract}
We show that some common and important 
global constraints like \alldiff and \gcc 
can be decomposed into simple arithmetic 
constraints on which we achieve bound 
or range consistency, and in some cases
even greater pruning. 
These decompositions can be easily
added to new solvers. 
They also provide other constraints
with access to the state of the propagator by
sharing 
of variables. Such sharing
can be used to improve propagation between constraints. 
We report experiments with our decomposition
in a pseudo-Boolean solver. 
\end{abstract}

\section{Introduction}

Global constraints allow users to specify
patterns that commonly occur in 
problems. One of the oldest and most useful
is the \alldiff constraint \cite{alice,regin1}.
This ensures that a set of variables are pairwise 
different. 
Global constraints can often
be decomposed into more primitive
constraints. For example, 
the \alldiff constraint can be decomposed
into a clique of binary inequalities. 
However, such decompositions
usually 
do not provide a 
global view and are thus not able to achieve 
levels of local
consistency, such as bound and domain consistency. 
Considerable
effort has therefore been invested in developing
efficient propagation algorithms to 
reason globally about such constraints. For instance,
several different propagation algorithms
have been developed for the \alldiff constraint
\cite{regin1,Leconte,puget98,mtcp2000,lopez1}.
In this paper, we show that 
several important global constraints
including \alldiff can be decomposed
into simple arithmetic constraints whilst
still providing a global view since  bound consistency can be
achieved.  

There are many reasons why such decompositions
are interesting. First, it is very surprising
that complex propagation algorithms can be
simulated by simple decompositions. In many
cases, we show that reasoning with the decompositions
is of similar complexity to 
existing monolithic propagation algorithms. 
Second, these decompositions can 
be easily added to a new solver. 
For example, we report experiments here
using these decompositions in a state of the art pseudo-Boolean
solver. 
We could just as easily use them in an ILP solver.
Third, introduced variables in these decompositions
give access to the state of the propagator. 
Sharing of such variables between 
decompositions can increase propagation. 
Fourth, these decomposition provide
a fresh perspective to propagating
global constraints that may be useful. 
For instance, our decompositions of
the \alldiff constraint suggest learning 
nogoods based on small Hall intervals. 

\section{Formal Background}

A constraint satisfaction problem (CSP) consists of a set of
variables, each with a finite domain of values, and a set of
constraints specifying allowed combinations of values for some
subset of variables. We use capitals for variables 
and lower case for values. 
We write $dom(X)$ for the domain of possible values for $X$, $min(X)$
for the smallest value in $dom(X)$,  $max(X)$ for the greatest,
and $range(X)$ for the interval $[min(X),max(X)]$.  
A \emph{global constraint} is one in which the number of variables $n$
is a parameter. For instance,
the global $\alldiff([X_1,\ldots,X_n])$ 
constraint ensures that 
$X_i \neq X_j$ for any $i<j$ \cite{regin1}.
We will 
assume values
range over 1 to $d$. 

Constraint solvers typically use
backtracking search to explore the space
of partial assignments. After each assignment,
propagation algorithms prune the search
space by enforcing local consistency properties like 
domain or bound consistency. A constraint is 
\emph{domain consistent} (\emph{DC})
iff when a variable is assigned any of the values in its domain, there
exist compatible values in the domains of all the other variables of
the constraint. Such an assignment is called
a \emph{support}. 
A constraint is \emph{bound consistent} (\emph{BC})
iff when a variable is
assigned the minimum or maximum value in its domain, there exist compatible
values between the minimum and maximum domain value
for all the other variables.
Such an assignment is called
a \emph{bound support}. 
Finally, between domain and bound consistency
is range consistency. A 
constraint is \emph{range consistent} (\emph{RC})
iff when a variable is
assigned any  value in its domain, there exists
a bound support. 

Constraint solvers usually  enforce local consistency after each
assignment down any branch in the search tree. For this reason, it is
meaningful to compute the total amortised cost of enforcing a
local consistency down an entire branch of the search tree so as to
capture the incremental cost of propagation.  
We will compute complexities in this way. 

\section{\alldiff constraint}\label{sec:alldiff}

The \alldiff constraint is one of the most useful global constraints
available to the constraint programmer. 
For instance, it can be used to specify that
activities sharing the same resource take place
at different times. 
A central concept in propagating
the \alldiff constraint
is the notion of a {\em Hall interval}.
This is an interval of 
$m$ domain values which completely contains 
the domains of $m$ variables. $[a,b]$ is 
a Hall interval iff 
$|\{ i \ | \ dom(X_i) \subseteq [a,b]\}|=b-a+1$. 
In any bound support, 
the variables whose domains are contained within
the Hall interval consume all the values
in the Hall interval, whilst any other variables
must find their support outside the Hall interval. 

\begin{myexample}
Consider an \alldiff constraint over
the following variables and values:
$$
{\scriptsize
\begin{array}{c|ccccc} 
 & 1 & 2 & 3 & 4 & 5 \\ \hline
X_1 & & & \ast & \ast & \\ 
X_2 & \ast& \ast & \ast & \ast & \\ 
X_3 & &  & \ast & \ast & \\ 
X_4 & & \ast & \ast & \ast & \ast \\
X_5 & \ast&  &  & &  
\end{array}
}
$$

$[1,1]$ is a Hall interval of size 1
as the domain of 1 variable, $X_5$ is  
completely contained within it. 
Therefore we can remove $[1,1]$
from the domains of all the other variables.
This leaves $X_2$ with a domain containing values 2,3,4. 

$[3,4]$ is a Hall interval
of size 2 as it completely contains
the domains of 2 variables, $X_1$ and $X_3$. We can
thus remove $[3,4]$ from the domains of $X_2$ and $X_4$. 
This leaves the following range  consistent
domains:
$$
{\scriptsize
\begin{array}{c|ccccc}
 & 1 & 2 & 3 & 4 & 5 \\ \hline
X_1 & & & \ast & \ast & \\ 
X_2 & & \ast & & & \\ 
X_3 & &  & \ast & \ast & \\ 
X_4 & & \ast& & & \ast\\ 
X_5 & \ast& & & &  
\end{array}
}
$$
Enforcing bound consistency on the same problem does not create
holes in domains. That is, it would leave  $X_4$ with the values
2,3,4,5.   
\end{myexample}

To identify and prune such Hall intervals
from the domains of other variables, 
Leconte has proposed a RC
propagator for the \alldiff constraint
\cite{Leconte} that runs in $\Theta(n^2)$ time. 
We now propose a simple decomposition of the \alldiff
constraint which permits us to enforce RC. 
The decomposition ensures that no interval
can contain more variables than its size. 
We introduce $O(nd^2)$ new 0/1 variables, $A_{ilu}$ to represent whether
$X_i$ takes a value in the interval $[l,u]$. 
For $1 \leq i \leq n$, $1 \leq l \leq u \leq d$ and $u-l < n$, 
we post the following constraints: 
\begin{eqnarray}
  A_{ilu}  = 1 &\iff & X_i \in [l, u] 
\label{eqn::firstRange} \label{eqn::secondRange} \\
  \sum_{i=1}^n A_{ilu} & \leq & u - l + 1 
\label{eqn::lastRange}
\end{eqnarray}
We illustrate this decomposition on our running example. 
\begin{myexample}
Consider again the last example (i.e. an \alldiff
constraint on $X_1 \in [3,4]$, $X_2 \in [1,4]$,
$X_3 \in [3,4]$,  $X_4 \in [2,5]$ and  $X_5 \in [1,1]$). 

First take the interval $[1,1]$.
Since $X_5 \in [1,1]$, (\ref{eqn::firstRange}) implies
$A_{511}=1$. 
Now from (\ref{eqn::lastRange}), 
$\sum_{i=1}^4 A_{i11} \leq 1$. 
That is, at most
one variable can take a value within
this interval. This means that
$A_{211}=0$. 
Using (\ref{eqn::firstRange}) and $A_{211}=0$, 
we get $X_2 \not\in [1,1]$. Since $X_2 \in [1,4]$,
this leaves $X_2\in [2,4]$. 

Now take the interval 
$[3,4]$.  From (\ref{eqn::firstRange}),
$A_{134}=A_{334}=1$. 
Now from (\ref{eqn::lastRange}), 
$\sum_{i=1}^4 A_{i34} \leq 2$. 
That is, at most
2 variables can take a value within
this interval. This means that 
$A_{234}=A_{434}=0$. 
Using (\ref{eqn::firstRange})
we get $X_2 \not\in [3,4]$, $X_4 \not\in [3,4]$. Since $X_2\in [2,4]$
and $X_4 \in [2,5]$,
this leaves $X_2=2$ and $X_4\in\{2,5\}$. Local reasoning
about the decomposition has thus made the
original \alldiff constraint range consistent. 
\end{myexample}

We will prove that enforcing DC on the decomposition
enforces RC on the original \alldiff constraint. 
We find it surprising that a simple decomposition like this
can simulate a complex propagation algorithm like Leconte's. 
In addition, the overall complexity of reasoning
with the decomposition is similar to Leconte's
propagator. 

\begin{mytheorem}
  \label{theorem::range}
  Enforcing 
  DC on constraints (\ref{eqn::secondRange})
  and (\ref{eqn::lastRange}) enforces 
  RC on the
  corresponding \alldiff constraint in $O(nd^3)$ down any
  branch of the search tree.
\end{mytheorem}
\myproof
\cite{Leconte} provides a necessary and sufficient
condition for RC of the \alldiff constraint:
every Hall interval should be removed from the domain
of variables whose domains are not fully
contained within that Hall interval. 
  Let $[a, b]$ be a Hall interval. That is, $|H| = b
  -a + 1$ where $H = \{i \mid dom(X_i) \subseteq [a,
  b]\}$. Constraint (\ref{eqn::secondRange}) fixes $A_{iab} = 1$ for
  all $i \in H$. The inequality (\ref{eqn::lastRange}) with $l = a$
  and $u = b$ becomes tight fixing $A_{iab} = 0$ for all $i \not\in
  H$. Constraint (\ref{eqn::secondRange}) for $l=a$, $u=b$, and $i
  \not\in H$ removes the interval $[a, b]$ from the domain of $X_i$
  as required for RC. 

  There are $O(nd^2)$ constraints (\ref{eqn::secondRange}) that can be
  woken $O(d)$ times down the branch of the search tree. 
Each
  propagation requires $O(1)$ time. Constraints
  (\ref{eqn::secondRange}) therefore take $O(nd^3)$ down the branch
  of the search tree to propagate. 
There are $O(d^2)$ constraints
  (\ref{eqn::lastRange}) that each take $O(n)$ time to propagate down
  the branch of the search tree for a total of $O(nd^2)$
  time. 
The total running time is 
given by  
$O(nd^3) + O(nd^2) = O(nd^3)$. 
\myqed

Note that if we use a solver in which
we can specify that constraints only wake
on reaching a particular bound, we can decrease  to $O(1)$ the
number of times  constraints (\ref{eqn::secondRange}) are woken, which
gives a total complexity in $O(nd^2)$. 

What about bound consistency of the \alldiff constraint? 
By using a representation that can only prune
bounds \cite{osccp07}, we can give a decomposition
that achieves BC in a similar way. 
In addition, we can reduce the overall complexity in the case
that constraints are woken whenever their bounds change. 
We introduce new 0/1 variables,
$B_{ik}$, $1 \leq k \leq d$  
and replace (\ref{eqn::secondRange})
by the following constraints:
\begin{eqnarray}
B_{il}=1 & \iff & X_i \leq l \label{eqn::secondBC} \\
A_{ilu}= 1 & \iff & (B_{i(l-1)}=0 \wedge B_{iu}=1) \label{eqn::thirdBC}
\end{eqnarray}

\begin{mytheorem}\label{theo:alldiffBC}
  Enforcing 
  BC on constraints 
 (\ref{eqn::lastRange}) to (\ref{eqn::thirdBC}) enforces 
  BC 
  on the corresponding \alldiff constraint
  in $O(nd^2)$ down any branch of the search tree.
\end{mytheorem}
\myproof
We first observe that BC is equivalent to DC on constraints
(\ref{eqn::lastRange}) because $A_{ilu}$ are Boolean variables. 
So, the proof follows that for Theorem~\ref{theorem::range} except
  that fixing $A_{ilu}=0$ prunes the bounds of $dom(X_i)$ if
  and only if $B_{i(l-1)}=0$ or $B_{iu}=1$, that is, if and only if exactly one
  bound of the domain of $X_i$ intersects the 
  interval $[l, u]$. Only the bounds that do not have a bound
  support are shrunk. 
The complexity reduces as 
(\ref{eqn::secondBC})
appears $O(nd)$ times and is woken $O(d)$ times, 
whilst (\ref{eqn::thirdBC}) appears $O(nd^2)$ times and is 
woken just $O(1)$ time.
\myqed

\myOmit{
As before, if our model already includes
the 0/1 variables $A_{ilu}$, we can 
enforce BC in $O(n^2d)$ time
down a branch of the search tree. 
We contrast this to Puget's original monolithic 
propagator, which is also based on Hall intervals, 
that takes $O(n \log n)$ time to enforce bound consistency
each time it is called \cite{puget98}. However,
as this is not incremental, it may be called
$O(nd)$ times down a branch. This gives a total
complexity of $O(n^2 d \log n)$ time down any
branch of the search tree. }

A special case of \alldiff is \permutation when we have
the same number of values as variables,
and the values are ordered consecutively. 
A decomposition of \permutation just needs
to replace (\ref{eqn::lastRange})
with the following equality
where (as before) $1 \leq l \leq u \leq d$, and $u - l < n$:
\begin{eqnarray}
  \sum_{i=1}^n A_{ilu} & = & u - l + 1 
\label{eqn::lastRangeEquality}
\end{eqnarray}
This can increase propagation. In some cases,
DC on constraints  (\ref{eqn::firstRange}) and
(\ref{eqn::lastRangeEquality})  will prune values that a RC
propagator for \permutation would miss. 
\begin{myexample} \label{permutation-example}
Consider a \permutation constraint over
the following variables and values:
$$
{\scriptsize
\begin{array}{c|ccc} 
 & 1 & 2 & 3 \\ \hline
X_1 & \ast & & \ast  \\ 
X_2 & \ast & & \ast \\ 
X_3 & \ast & \ast & \ast 
\end{array}
}
$$
These domains are range consistent. 
However, take the interval $[2,2]$. 
By DC on (\ref{eqn::firstRange}),
$A_{122}=A_{222}=0$. Now, from 
(\ref{eqn::lastRangeEquality}), 
we have $\sum_{i=1}^3 A_{i22}=1$. Thus
$A_{322}=1$. By (\ref{eqn::firstRange}),
this sets $X_3=2$. On this particular problem instance, DC on   constraints  (\ref{eqn::firstRange}) and
(\ref{eqn::lastRangeEquality}) has enforced domain consistency
on the original \alldiff constraint. 
\end{myexample}

\section{\gcc constraint}

A generalization of the \alldiff constraint
is the global cardinality constraint, 
$\gcc([X_1,\ldots, X_n],[l_1,\ldots,l_m],[u_1,\ldots,u_m])$.
This ensures that the value $i$ occurs
between $l_i$ and $u_i$ times in $X_1$ to
$X_n$. The \gcc constraint is useful 
in resource allocation problems where values
represent resources. For instance, in the
car sequencing problem (prob001 at CSPLib.org),
we can post a \gcc constraint to ensure that
the correct number of cars of each type
is put on the assembly line. 

We can decompose
\gcc in a similar way to \alldiff 
but with an additional $O(d^2)$ 
integer variables, $N_{lu}$ to represent the number
of variables using values in each interval $[l,u]$. 
Clearly, $N_{lu} \in [\sum_{i=l}^u l_i, \sum_{i=l}^u u_i]$
and $N_{1d}=n$. 
We then post the following constraints
for $1 \leq i \leq n$, $1 \leq l \leq u \leq d$, 
$1 \leq k < u$:
\begin{eqnarray}
  A_{ilu} = 1 & \iff & 
X_i \in [l,u] \label{eqn::firstGCC} \\
  N_{lu} & = & \sum_{i=1}^n A_{ilu}    \label{eqn::bigAGCC}\\
  N_{1u} & = & N_{1k} + N_{(k+1)u} \label{eqn::GCCTriangle} \label{eqn::lastGCC} 
\end{eqnarray}

\begin{myexample} 
Consider a \gcc constraint with the following variables
and upper and lower bounds on the occurrences of values: 
$$
{\scriptsize
\begin{array}{c|ccccc} 
v & 1 & 2 & 3 & 4& 5\\ \hline
X_1 & \ast& & & &\\ 
X_2 &\ast & \ast &\ast & \ast&\ast\\ 
X_3 & &  &\ast  &  &\\
X_4 &\ast & \ast &\ast & \ast&\ast\\ 
X_5 &\ast & \ast &\ast & \ast&\ast\\ 
\hline
l_v & 1 & 1 & 0 & 1&1 \\
u_v & 5 & 5 & 5 & 5&5
\end{array}
}
$$
Enforcing RC 
removes 1 and 3 from  $X_2,X_4$ and $X_5$ and leaves
the other domains unchanged. 
We can derive this from our decomposition. From 
the lower and upper bounds
on the number of occurrences of the values, we have $N_{ii} \in [1,5]$
except for $N_{33}\in [0,5]$ and we have 
$N_{12}\in [2,10], N_{13}\in [2,15]$ and $N_{14}\in[3,20]$. 
By (\ref{eqn::firstGCC}), 
$A_{333}=1$. From (\ref{eqn::bigAGCC}),
$N_{33}=\sum_{i=1}^6 A_{i33} \in [1,5]$.  From 
$N_{15}=N_{14} + N_{55}$ we have $N_{14}\in[3,4]$ (i.e., upper bound decreased)
because  
$N_{15}=5$ and  $N_{55} \in [1,5]$. Similarly, we derive from
$N_{14}=N_{13} + N_{44}$ that $N_{13}\in[2,3]$ and from   
$N_{13}=N_{12} + N_{33}$ that  $N_{12}\in[2,2]$. 
 From the same constraint, we shrink $N_{13}$ to $[3,3]$ and $N_{33}$
to $[1,1]$. 
Finally, $N_{12}=N_{11} + N_{22}$ shrinks   $N_{11}$ to  $[1,1]$. 
By (\ref{eqn::firstGCC}), $A_{111}= A_{333}=1$, so by (\ref{eqn::bigAGCC}),
$A_{i11}=0, i, \in 2..5$ and $A_{i33}=0, i, \in \{1,2,4,5\}$. 
By (\ref{eqn::firstGCC}), this removes 1 and 3 from $X_2,X_4,X_5$. 
Local reasoning about the decomposition has 
made the original \gcc constraint range consistent.
\end{myexample}

\myOmit{
\begin{myexample} Taken from \cite{bc-gcc}. 
Consider a \gcc constraint with the following variables
and upper and lower bounds on the occurrences of values: \cb{Can we find a simple example which
uses the triangle?}
$$
{\scriptsize
\begin{array}{c|cccc} 
v & 1 & 2 & 3 & 4\\ \hline
X_1 & & \ast & & \\ 
X_2 & \ast & \ast & & \\ 
X_3 & & \ast & \ast &  \\
X_4 & & \ast & \ast &  \\
X_5 & \ast & \ast & \ast & \ast \\
X_6 & & & \ast & \ast   \\ \hline
l_v & 1 & 1 & 1 & 2 \\
u_v & 3 & 3 & 3 & 3
\end{array}
}
$$
Enforcing BC 
sets $X_2=1$, $X_5=X_6=4$ and leaves
the other domains unchanged. 
We can derive this from our decomposition. From 
the lower and upper bounds
on the number of occurrences of the value
4, we have $N_{44} \in [2,3]$.
By (\ref{eqn::firstGCC}), 
$A_{i44}=0$ for $i\leq 4$. 
But, by (\ref{eqn::bigAGCC}),
$\sum_{i=1}^6 A_{i44} \in [2,3]$. 
This means that $A_{544}=A_{644}=1$. 
By (\ref{eqn::firstGCC}), this
sets $X_5=X_6=4$. 
Similarly, we have $N_{11} \in [1,3]$.
By (\ref{eqn::firstGCC}), 
$A_{i11}=0$ for $i\neq 2$. 
But, by (\ref{eqn::bigAGCC}),
$\sum_{i=1}^6 A_{i11} \in [1,3]$. 
This means that $A_{211}=1$. 
By (\ref{eqn::firstGCC}), this
sets $X_2=1$. 
\end{myexample}
}

We next show that enforcing DC on 
constraint~(\ref{eqn::firstGCC}) and BC on constraints (\ref{eqn::bigAGCC}) and (\ref{eqn::lastGCC}) enforces 
RC on the \gcc constraint. 

\begin{mytheorem}
  \label{thm:gcc-bc}
  Enforcing DC 
  on constraint
  (\ref{eqn::firstGCC}) and BC on constraints (\ref{eqn::bigAGCC}) and (\ref{eqn::lastGCC}) achieves RC 
  on the corresponding \gcc constraint in
  $O(nd^3)$ time down any
  branch of the search tree.
\end{mytheorem}
\myproof
We use $I_V$ 
for the
  number of variables $X_i$ whose 
range
  $range(X_i)$
intersects the set $V$ of values, 
and $S_V$ 
for the
  number of variables $X_i$ whose 
range
is a subset of $V$. 
We first show that if RC fails on the \gcc, DC on (\ref{eqn::firstGCC}) and BC on (\ref{eqn::bigAGCC}) and (\ref{eqn::lastGCC})
will fail. 
We derive from \cite[Lemmas 1 and 2]{QuimperGLB05} that 
RC
fails  on a \gcc  if and only if there exists a set of
values $V$ such that 
$S_V>
\sum_{v\in V}u_{v}$ 
or  
such that 
$I_V<\sum_{v\in V}l_{v}$. 
Suppose first a set $V$ such that 
$S^{}_V>
\sum_{v\in V}u_{v}$. The fact that domains are considered as intervals
implies that either $range(V)$ includes more variable domains than
the sum of the upper bounds  (like $V$), or the union of the $range(X_i)$ that are
included in $V$ lets a hole of unused values in $V$, which implies that
there exists an interval $[l,u]\subset V$ such that 
$S^{}_{[l,u]}>\sum_{v\in [l,u]}u_{v}$. 
So, in any case,  there exists an interval $[l,u]$ in  $V$ with 
$S^{}_{[l,u]}>\sum_{v\in [l,u]}u_v$. 
By 
(\ref{eqn::firstGCC}) we
have $\sum_{i=1}^n A_{ilu}\geq S^{}_{[l,u]}$ whereas the greatest
value in the domain of
$N_{lu}$ was set to $\sum_{v\in [l,u]}u_v$. 
So BC will fail on 
$N_{lu}  =  \sum_{i=1}^n A_{ilu}$. 
Suppose now that a set $V=\{v_{1}, \ldots, v_{k}\}$ is such that 
$I^{}_V<\sum_{v_i\in V}l_{v_i}$. 
The total number of values taken by $X_i$ variables 
being equal to $n$, the number of
variables $X_i$ with $range(X_i)$ not intersecting $V$ is greater than
$n-\sum_{v_i\in V}l_{v_i}$, that is,
$  S^{}_{[1,v_1-1]} +S^{}_{[v_1+1,v_2-1]} 
+  \ldots +S^{}_{[v_{k}+1,d]}
>n-\sum_{v_i\in V}l_{v_i}$. 
Thanks to 
(\ref{eqn::bigAGCC}), we
know that for any $l,u$, 
 $N_{lu} \geq S^{}_{[l,u]} $. 
So, $N_{1(v_1-1)} +N_{(v_1+1)(v_2-1)} 
+  \ldots +N_{(v_{k}+1)d}
>n-\sum_{v_i\in V}l_{v_i}$. 
The initial domains of $N_{lu}$ variables  
also tell us that for every $v_i$ in $V$,  
$N_{v_i v_i}\geq l_{v_i}$. 
Thus, 
$min(N_{1(v_1-1)}) +min(N_{v_1v_1}) +
  min(N_{(v_1+1)(v_2-1)}) + \ldots +
  min(N_{(v_{k}+1)d})>n=max(N_{1d})$. Successively applying BC on 
 $N_{1v_1}=N_{1(v_1-1)} +N_{v_1v_1}$, then on 
$N_{1(v_2-1)}=N_{1v_1} +N_{(v_1+1)(v_2-1)}$, and so on until 
$N_{1d}=N_{1v_k} +N_{(v_k+1)d}$ will successively increase the
  minimum 
of these variables and will lead to a failure on
  $N_{1d}$. 
%

We now show that when DC on (\ref{eqn::firstGCC}) and BC on (\ref{eqn::bigAGCC}) and (\ref{eqn::lastGCC})
do not fail, it 
prunes all values that are pruned when enforcing RC on the \gcc constraint. 
  Consider a value $v\in dom(X_q)$ for some $q\in 1..n$ such that
  $v$ does not have any bound support. 
We derive from
  \cite[Lemmas 1 and 6]{QuimperGLB05} that 
a value $v$ for a  variable $X_q$ does not have a bound support on \gcc if
and only if 
there exists 
a set $ V$ of values such that either (i) 
$S^{}_{V}=\sum_{w\in V}u_w$, $v\in V$ and  $range(X_q)$ is not
included in $ V$, 
or  (ii) 
$I^{}_{V}=\sum_{w\in V}l_w$,  $v\notin V$ and $range(X_q)$
intersects $V$. 
In case (i),  $V$  contains $v$ and the values it contains   
  will be taken by too many variables if $X_q$ is in it. 
In case (ii),  $V$ does  not contain $v$ and its values will be
  taken by not enough  variables if $X_q$ is not in it. 
Consider case (i): Since DC did not fail on (\ref{eqn::firstGCC}), 
by a similar reasoning as above for detecting 
failure, we derive that $V$ is composed of intervals  $[l,u]$ such that 
$S^{}_{[l,u]}=\sum_{w\in [l,u]}u_w$. Consider the interval
$[l,u]$ containing $v$. 
The greatest value in the  initial domain of $N_{lu}$ was
$\sum_{w\in [l,u]}u_w$, which is exactly the number of variables
with range included in $[l,u]$ without counting $X_q$ because its
range is not included in $V$. Thus, 
(\ref{eqn::bigAGCC}) forces 
  $A_{qlu}=0$ and 
(\ref{eqn::firstGCC}) 
 prunes value  $v$ from $dom(X_q)$ because 
$v\in [l,u]$ by assumption. 
Consider now case (ii): $V=\{v_{1}, \ldots, v_{k}\}$ is such that
  $I^{}_V=\sum_{v_i\in V}l_{v_i}$.  The total number of values taken by
  the $X_i$ variables being equal to $n$, the number of variables
  $X_i$ with $range(X_i)$ not intersecting $V$ is equal to
  $n-\sum_{v_i\in V}l_{v_i}$, that is $ S^{}_{[1,v_1-1]}
  +S^{}_{[v_1+1,v_2-1]} + \ldots +S^{}_{[v_{k+1},d]} =n-\sum_{v_i\in
    V}l_{v_i}$.  Thanks to 
  (\ref{eqn::bigAGCC}), we know that for any $l,u$, $N_{lu} \geq
  S^{}_{[l,u]} $.  So, $N_{1(v_1-1)} +N_{(v_1+1)(v_2-1)} + \ldots
  +N_{(v_{k+1})d} \geq n-\sum_{v_i\in V}l_{v_i}$.  The initial domains of
  $N_{lu}$ variables also tell us that for every $v_i$ in $V$, $N_{v_i
    v_i}\geq l_{v_i}$. 
%
%
 Thus, $min(N_{1(v_1-1)}) + min(N_{v_1v_1}) +
   min(N_{(v_1+1)(v_2-1)}) + \ldots + min(N_{(v_{k}+1)d})\geq
   n=max(N_{1d})$.  Successively applying BC on $N_{1v_1}=N_{1(v_1-1)}
   +N_{v_1v_1}$, then on $N_{1(v_2-1)}=N_{1v_1} + N_{(v_1+1)(v_2-1)}$,
   and so on until $N_{1v_k}=N_{1(v_k-1)} + N_{v_kv_k}$ will increase
   all $min(N_{1(v_i-1)})$ and $min(N_{1v_i})$, to the sum of the
   minimum values of the variables in the right side of each constraint
   so that $min(N_{1v_k})= min(N_{1(v_1-1)}) +min(N_{v_1v_1}) +
   min(N_{(v_1+1)(v_2-1)}) + \ldots + min(N_{v_{k}v_k})$.  Then,
   because $max(N_{1d})=n$, BC on $N_{1d}=N_{1v_k} +N_{(v_k+1)d}$ will
   decrease the maximum value of $N_{1v_k}$ and $N_{(v_k+1)d}$ to their
   minimum value, BC on $N_{1v_k}=N_{1(v_k-1)} +N_{v_kv_k}$ will
   decrease the maximum value of $N_{1(v_k-1)}$ and $N_{v_kv_k}$ to
   their minimum value, and so on until all $N_{(v_i+1)(v_{i+1}-1)}$
   are forced to the singleton
   $min(N_{(v_i+1)(v_{i+1}-1)})=S^{}_{[v_i+1,v_{i+1}-1]} $.
At this point, 
(\ref{eqn::bigAGCC})
 forces $A_{j(v_i+1)(v_{i+1}-1)}=0$ for every variable $X_j$ with range
 not included in  the interval $[v_i+1,v_{i+1}-1]$ because that
 interval is saturated by variables $X_p$ in
 $S^{}_{[v_i+1,v_{i+1}-1]}$, for which  $A_{p(v_i+1)(v_{i+1}-1)}=1$. 
By assumption value $v$ is not in $V$, 
so there exists such an
 interval $[v_i+1,v_{i+1}-1]$ that contains $v$. Furthermore,
 $range(X_q)$ intersects $V$, so it is not included in 
$[v_i+1,v_{i+1}-1]$. 
Therefore,  
$A_{q(v_i+1)(v_{i+1}-1)}$ is forced to 0 and 
 (\ref{eqn::firstGCC})  
 prunes $v$ from $dom(X_q)$.

  There are $O(nd^2)$ constraints (\ref{eqn::firstGCC}) that can be
  woken $O(d)$ times down the branch of the search tree in  $O(1)$, so
  a  total of $O(nd^3)$
  down the branch. 
  There are $O(d^2)$ constraints
  (\ref{eqn::bigAGCC}) which can be woken $O(n)$ times each down the
  branch 
for a total cost in  $O(n)$ time down the
  branch. Thus  a  total of $O(nd^2)$.  
There are $O(d^2)$ constraints (\ref{eqn::GCCTriangle}) that can
  be woken $O(n)$ times down the branch. 
Each
  propagation takes $O(1)$ time to execute for a total of $O(nd^2)$
  time down the branch. 
  The final complexity down
  the branch of the search tree is therefore $O(nd^3) + O(nd^2) +
  O(nd^2) = O(nd^3)$. 
\myqed


What about bound consistency of the \gcc constraint? 
As in the case of \alldiff, by replacing constraints
(\ref{eqn::firstGCC}) by constraints (\ref{eqn::secondBC}) and
(\ref{eqn::thirdBC}), 
the decomposition  achieves BC. 

\begin{mytheorem}
  \label{thm:gcc-bc2}
  Enforcing BC 
  on  constraints (\ref{eqn::secondBC}), 
  (\ref{eqn::thirdBC}), 
  (\ref{eqn::bigAGCC}) and (\ref{eqn::lastGCC}) achieves BC 
  on the corresponding \gcc constraint in
  $O(nd^2)$ time down any
  branch of the search tree.
\end{mytheorem}
\myproof
The proof follows that for Theorem~\ref{thm:gcc-bc} except
  that fixing $A_{ilu}=0$ prunes the bounds of $dom(X_i)$ if
  and only if  exactly one
  bound of the domain of $X_i$ intersects the 
  interval $[l, u]$. 
The complexity reduces to $O(nd^2)$ as BC on 
(\ref{eqn::secondBC})
 and  (\ref{eqn::thirdBC}) is in  $O(nd^2)$  (see Theorem
 \ref{theo:alldiffBC}) and BC on  (\ref{eqn::bigAGCC}) and (\ref{eqn::lastGCC}) is in  $O(nd^2)$  (see Theorem
 \ref{thm:gcc-bc}). 
\myqed

The best known   algorithm for BC on \gcc runs in  
$O(n)$ time at each call \cite{QuimperGLB05} and can be  
awaken $O(nd)$ times down a  branch
. This gives
a total of $O(n^2d)$, which is greater than the  $O(nd^2)$ here
when $n>d$. 
Our decomposition is also interesting because, as we show
in the next section, we can 
use it to combine together propagators.

\section{Other global constraints}

Many other global constraints that 
count variables or values can be decomposed
in a similar way. 
%
%
For example, the global
constraint $\mypermutation([X_1, \ldots, X_n], [Y_1, \ldots, Y_n])$ is
satisfied if and only if the $Y_i$ variables are a permutation of the
$X_i$ variables. A monolithic flow-based propagator
for this constraint 
is given in \cite{bktcpaior04}. 
The following decomposition encodes the \mypermutation\
constraint where 
$1 \leq i \leq n$, $1 \leq l \leq u \leq d$, 
$l \leq k$ and $k < u$:
\begin{eqnarray*}
&  A_{ilu} = 1 \iff X_i \in [l,u], &
  B_{ilu} = 1 \iff Y_i \in [l,u] \\
&  N_{lu} = \sum_{i=1}^n A_{ilu}, & 
  N_{lu} = \sum_{i=1}^n B_{ilu} \\
&  N_{1u} = N_{1k} + N_{(k+1)u} &
\end{eqnarray*}

This decomposition can be obtained by posting
decompositions for 
$\egcc([X_1,\ldots,X_n],[O_1,\ldots,O_m])$ and
$\egcc([Y_1,\ldots,Y_n],[O_1,\ldots,O_m])$
and eliminating common sub-expressions
(\egcc\ is an extended form of the \gcc constraint
in which upper and lower bounds on occurrences of
values are replaced by integer variables). 
This is another argument in favor of decompositions since
it allows constraints to share ``internal''
state through common intermediate variables. 
Such sharing
can increase propagation.

\myOmit
{
\begin{mytheorem}
  Enforcing BC on the above constraints
  is strictly stronger than 
  BC on $\gcc([X_1,\ldots,X_n],[O_1,\ldots,O_m])$ and
  on $\gcc([Y_1,\ldots,Y_n],[O_1,\ldots,O_m])$.
\end{mytheorem}
\myproof  By Theorem \ref{thm:gcc-bc}
it is clearly at least as strong.
To show strictness, consider the following 
example:
$$
{\scriptsize
\begin{array}{c|ccccc} 
 & 1 & 2 & 3 & 4 & 5\\ \hline
X_1 & \ast &  \ast &  \\ 
X_2 &      &       & \ast & \ast & \ast \\ 
Y_1 & \ast & \ast & \ast \\
Y_2 &      &      &       & \ast & \ast
\end{array}
}
$$ 
If we have $O_i \in [0, 1]$ for $1 \leq i \leq 5$ then both
$\egcc([X_1,X_2],[O_1,O_2,O_3,O_4,O_5])$ and
$\egcc([Y_1,Y_2],[O_1,O_2,O_3,O_4,O_5])$ are BC. However, enforcing BC on
the decomposition removes $3$ from the domain of $X_2$ and $Y_1$.
\myqed
}

\begin{myexample}
Consider the following 
example:
$$
{\scriptsize
\begin{array}{c|ccccc} 
 & 1 & 2 & 3 & 4 & 5\\ \hline
X_1 & \ast &  \ast &  \\ 
X_2 &      &       & \ast & \ast & \ast \\ 
Y_1 & \ast & \ast & \ast \\
Y_2 &      &      &       & \ast & \ast
\end{array}
}
$$ 
If we have $O_i \in [0, 1]$ for $1 \leq i \leq 5$ then both
$\egcc([X_1,X_2],[O_1,O_2,O_3,O_4,O_5])$ and
$\egcc([Y_1,Y_2],[O_1,O_2,O_3,O_4,O_5])$ are BC. However, enforcing BC on
the decomposition of 
$\mypermutation([X_1, X_2], [Y_1, Y_2])$ 
removes $3$ from the domain of $X_2$ and $Y_1$.
\end{myexample}
 
In fact, we conjecture that enforcing BC on this
decomposition achieves BC on 
the \mypermutation constraint itself. 
Similar decompositions can be given for other global constraints
like 
\nvalue 
and 
\common . 

\section{Experimental Results}

\newcommand{\csum}{\constraint{Sum}\xspace}
\newcommand{\mem}{\constraint{Member}\xspace}

To test these decompositions,
we ran experiments on pseudo-Boolean encodings (PB) 
of CSPs containing
\alldiff and \permutation 
constraints. We used the MiniSat+ 1.13 solver on an
Intel Xeon 4 CPU, 2.0 Ghz, 4G RAM with 
a timeout of $600$ seconds for each experiment. 
%
%
Our decompositions 
contain two types of constraints:  
\csum constraints like (\ref{eqn::lastRange})
and \mem constraints like (\ref{eqn::firstRange}). 
The \csum constraints  
is posted directly to the MiniSat+ solver. To encode \mem 
constraints, 
we use literals
$B_{ij}$ for the truth of $X_i \leq j$~\cite{osccp07},
and clauses of the form
$(A_{ilu} = 1) \Leftrightarrow ( B_{i(l-1)}=0 \wedge B_{iu}=1)$. 
This achieves bound consistency (Theorem
\ref{theo:alldiffBC}).  
To increase propagation, we 
use a direct encoding with literals 
$Z_{ij}$ for the truth of $X_i = j$ and
clauses 
$(A_{ilu}=0) \Rightarrow  (Z_{ij}=0)$, $j \in [l,u]$. The overall
consistency achieved is therefore between BC and RC. 
We denote this encoding $HI$.  
To explore the impact
of small Hall intervals, we also tried $HI_k$,  
a PB encoding 
with
only those constraints (\ref{eqn::lastRange})
for which $u - l + 1 \leq k$. This
detects Hall intervals of size at most $k$. 
Finally, we decomposed \alldiff 
into a clique of binary
inequalities, and used a direct encoding 
to convert this into SAT (denoted 
$BI$).

\paragraph{Pigeon Hole Problems.}

Table~\ref{t:t1} gives results on pigeon hole problems (PHP)
with $n$ pigeons and \mbox{$n-1$} holes. 
Our decomposition is both faster and gives 
a smaller search tree compared to the $BI$ decomposition. 
On such problems, 
detecting large Hall intervals is essential. 
\begin{table}[tb]
\begin{center}
 {\scriptsize
\begin{tabular}{|@{}c@{}||@{}r@{}@{}l@{}|@{}r@{}@{}l@{}|@{}r@{}@{}l@{}|@{}r@{}@{}l@{}|@{}r@{}@{}l@{}|@{}r@{}@{}l@{}|}
\hline
n & \multicolumn {2}{|c|}{$BI$} & 
 \multicolumn {2}{|c|}{$HI_{1}$} & 
 \multicolumn {2}{|c|}{$HI_{3}$} & 
 \multicolumn {2}{|c|}{$HI_{5}$} & 
 \multicolumn {2}{|c|}{$HI_{7}$} & 
 \multicolumn {2}{|c|}{$HI_{9}$} \\ 
& 
  $bt$ &/ $t$ & 
  $bt$ &/ $t$ & 
  $bt$ &/ $t$ & 
  $bt$ &/ $t$ & 
  $bt$ &/ $t$ & 
  $bt$ &/ $t$ \\ 
\hline 
\hline
5&        30 &/  \textbf{ 0.0}  
&        28 &/  \textbf{ 0.0}  
&       \textbf{  4} &/  \textbf{ 0.0}  
&   &    
&   &    
&   &    
\\ 
7&       622 &/  \textbf{ 0.0}  
&       539 &/  \textbf{ 0.0}  
&        47 &/  \textbf{ 0.0}  
&       \textbf{ 6} &/  \textbf{ 0.0}  
&   &    
&   &    
\\ 
9&    16735 &/   0.3  
&    18455 &/   0.7  
&       522 &/  \textbf{ 0.0}  
&       122 &/  \textbf{ 0.0}  
&     \textbf{ 8} &/  \textbf{ 0.0}  
&   &    
\\ 
11&   998927 &/  29.3  
&   665586 &/  44.8  
&     5681 &/   0.3  
&       171 &/  \textbf{ 0.0}  
&       180 &/  \textbf{ 0.0}  
&     \textbf{ 10} &/   0.1  
\\ 
13&  - &/  -  
&  - &/  -  
&    13876 &/   0.9  
&     2568 &/   0.2  
&       247 &/  \textbf{ 0.1}  
&       \textbf{ 195} &/  \textbf{ 0.1}  
\\ 
15&  - &/  -  
&  - &/  -  
&  1744765 &/ 188.6  
&    24109 &/   2.6  
&     1054 &/   0.2  
&      \textbf{ 165} &/  \textbf{ 0.1}  
\\ 
17&  - &/  -  
&  - &/  -  
&  - &/  -  
&   293762 &/  48.0  
&     8989 &/   1.1  
&    \textbf{ 4219} &/  \textbf{ 0.6}  
\\ 
19&  - &/  -  
&  - &/  -  
&  - &/  -  
&   107780 &/  21.8  
&   857175 &/ 368.0  
&    \textbf{ 39713} &/  \textbf{ 9.9}  
\\ 
21&  - &/  -  
&  - &/  -  
&  - &/  -  
&  - &/  -  
&   550312 &/ 426.2  
&    \textbf{ 57817} &/  \textbf{ 33.5}  
\\ 
\hline 
\end{tabular} 
 \caption{\label{t:t1}PHP problems. $t$ is time and $bt$ is the number of backtracks to solve the problem. }
} 
\end{center} 
 \end{table} 
 
 \paragraph{Double-Wheel Graceful Graphs.}

The second set of experiments uses double-wheel graceful 
graphs~\cite{pscp03}.
We converted the CSP model in~\cite{pscp03} into a PB formula.  
This model has an \alldiff constraint on node labels and 
a \permutation constraint on  edge labels. 
For the \permutation constraint we use (\ref{eqn::lastRangeEquality}). 
We strengthen the $BI$ decomposition with
clauses to ensure that every value appears at least once. 
Table~\ref{t:t2}
show that our decomposition outperforms 
the augmented $BI$ decomposition on many instances.
Whilst detecting large Hall intervals can greatly reduce search, 
in some cases the branching 
heuristics appear to be fooled by the 
extra variables introduced in the encodings. 

Overall these  experiments suggest that detecting Hall intervals reduces
search significantly, and focusing on small Hall intervals
may be best except on problems where large Hall intervals 
occur frequently. 

\begin{table}[tb]
\begin{center}
 {\scriptsize
\begin{tabular}{|@{}c@{}||@{}r@{}@{}l@{}|@{}r@{}@{}l@{}|@{}r@{}@{}l@{}|@{}r@{}@{}l@{}|@{}r@{}@{}l@{}|@{}r@{}@{}l@{}|}
\hline
$DW_n$& \multicolumn {2}{|c|}{$BI$} & 
 \multicolumn {2}{|c|}{$HI_{1}$} & 
 \multicolumn {2}{|c|}{$HI_{3}$} & 
 \multicolumn {2}{|c|}{$HI_{5}$} & 
 \multicolumn {2}{|c|}{$HI_{7}$} & 
 \multicolumn {2}{|c|}{$HI_{9}$} \\ 
& 
  $bt$ &/ $t$ & 
  $bt$ &/ $t$ & 
  $bt$ &/ $t$ & 
  $bt$ &/ $t$ & 
  $bt$ &/ $t$ & 
  $bt$ &/ $t$ \\ 
\hline 
\hline 
3&      176 &/   0.1  
&       90 &/   0.1  
&       \textbf{ 63} &/  \textbf{ 0.1}  
&   &    
&   &    
&   &    
\\ 
4&         30 &/  \textbf{ 0.1  }  
&       \textbf{14} &/   0.1  
&      212 &/   0.2  
&   &    
&   &    
&   &    
\\ 
5&        \textbf{ 22} &/  \textbf{ 0.2}  
&      526 &/   0.4  
&       87 &/   0.3  
&     1290 &/   1.7  
&   &    
&   &    
\\ 
6&     1341 &/   1.0  
&      873 &/   0.9  
&      \textbf{ 318} &/  \textbf{ 0.7  }  
&     1212 &/   2.9  
&   &    
&   &    
\\ 
7&     2948 &/   3.6  
&     2047 &/   4.2  
&     1710 &/   3.6  
&     1574 &/   4.0  
&        \textbf{ 27} &/  \textbf{ 0.9  }  
&   &    
\\ 
8&     2418 &/   5.5  
&       724 &/  \textbf{ 2.2  }  
&      643 &/   2.8  
&     \textbf{ 368} &/   2.4  
&     3955 &/  19.5  
&   &    
\\ 
9&     3378 &/   8.6  
&     1666 &/   5.7  
&     1616 &/   9.0  
&       \textbf{ 30} &/  \textbf{ 1.8 }  
&    10123 &/ 129.7  
&      405 &/   6.5  
\\ 
10&    19372 &/ 118.3  
&     9355 &/  66.2  
&    14120 &/  85.9  
&       \textbf{ 10} &/  \textbf{ 2.1}  
&     4051 &/  35.0  
&     5709 &/  71.2  
\\ 
11&      839 &/   5.4  
&    12356 &/  84.2  
&     1556 &/  13.9  
&        \textbf{ 14} &/  \textbf{ 2.4}  
&     7456 &/ 105.2  
&     5552 &/  92.7  
\\ 
\hline 
\end{tabular} 
 \caption{\label{t:t2}Double-wheel graceful graphs. $t$ is time and $bt$ is the number of backtracks to solve the problem }
} 
\end{center} 
 \end{table}

\section{Other Related Work}

The \alldiff constraint first appeared in the ALICE
constraint programming language \cite{alice}. 
Regin proposed a DC propagator 
that runs in $O(n^{2.5})$ time \cite{regin1}.
Leconte gave a RC propagator based on Hall intervals
that runs in $O(n^2)$ time 
\cite{Leconte}. Puget then developed a BC propagator
also based on Hall intervals that runs in $O(n \log(n))$ time \cite{puget98}.
This 
was later improved by Melhorn and Thiel \cite{mtcp2000}
and then Lopez-Ortiz {\it et al.} \cite{lopez1}. 

The global cardinality constraint, \gcc
was introduced in the CHARME 
language \cite{omtexpert89}. 
Regin proposed a DC propagator based on
network flow that runs in $O(n^2)$ time \cite{regin2}.
Katriel and Thiel proposed a BC propagator
for the \egcc constraint 
\cite{bc-gcc2}. 
Quimper {\it et al.} proved that enforcing DC on 
the \egcc constraint is NP-hard
\cite{qblgcp04}. They also improved the time
complexity to enforce DC and gave the first propagator
for enforcing RC on \gcc .

Many decompositions have been given for
a wide range of global constraint. However, 
decomposition in general tends to hinder propagation.
For instance, \cite{swijcai99} shows
that the decomposition of \alldiff constraints into binary
inequalities hinders propagation
. 
On the other hand, there are global constraints
where decompositions have been given 
that do not hinder propagation. For example,
Beldiceanu {\it et al.} identify conditions
under which global constraints specified
as automata can be decomposed into signature
and transition constraints without hindering
propagation \cite{mustadd}. 
As a second example, many global constraints
can be decomposed using \roots and \range
which can themselves often be 
propagated effectively 
using simple decompositions \cite{rootsrange,range,roots}. 
As a third example, 
decompositions of the \regular and \grammar constraints have
been given that do not hinder propagation
\cite{qwcp06,qwcp07,qwaaai08,bhhkwecai08,knwcpaior08}. 
As a fourth example,
decompositions of the \sequence constraint 
have been shown to be effective
\cite{bnqswcp07}. 
Finally,
the \precedence constraint
can be decomposed into ternary
constraints without hindering propagation
\cite{wecai06}.

\section{Conclusions}

We have shown that some common global
constraints like \alldiff and \gcc 
can be decomposed into simple arithmetic 
constraints whilst still maintaining a global
view that achieves range or bound consistency. 
These decompositions are interesting for a number
of reasons. First, we can easily incorporate them
into other solvers. 
Second, the decompositions provide other constraints
with access to the state of the propagator. 
Third, these decompositions provide
a fresh perspective on propagation of
global constraints. For instance, our results
suggest that
it may pay to focus propagation and nogood learning 
on small Hall intervals.
Finally, these decompositions raise an important question.
Are there propagation algorithms that cannot be
efficiently simulated using decompositions? In 
\cite{comics-cnf09}, we use circuit
complexity to argue that a domain consistency
propagator for the \alldiff
constraint 
cannot be simulated using a polynomial
sized decomposition.

\bibliographystyle{named}



\end{document}